\documentclass{llncs}
\usepackage{url}
\usepackage{paralist}
\usepackage{amsmath}
\usepackage{booktabs}
\usepackage{multirow}
\usepackage{lscape}
\usepackage[final]{graphicx}
\usepackage{array}
\usepackage[T1]{fontenc}
\usepackage{hyperref}

\usepackage[all]{hypcap}

\newcommand{\cnf}{\textsc{cnf}}
\newcommand{\sat}{\textsc{sat}}
\newcommand{\csp}{\textsc{csp}}

\newcommand{\false}{\emph{false}}
\newcommand{\satzilla}{\textsc{SATzilla}}

\newcommand{\mistral}{\texttt{mistral}}

\title{Transformation-based Feature\\Computation for Algorithm Portfolios}

\author{Barry Hurley$^1$ \and Serdar Kadioglu$^2$ \and Yuri Malitsky$^1$ \and Barry O'Sullivan$^1$}
\institute{Insight Centre for Data Analytics\\
Department of Computer Science, University College Cork, Ireland\\
\email{\{b.hurley|y.malitsky\}@4c.ucc.ie}, \email{b.osullivan@cs.ucc.ie} \and
Oracle America Inc., Burlington, MA 01803, USA \\
\email{serdark@cs.brown.edu}}

\begin{document}

\maketitle

\setcounter{footnote}{0}

\begin{abstract}

Instance-specific algorithm configuration and algorithm portfolios have been
shown to offer significant improvements over single algorithm approaches in a
variety of application domains. In the \sat\ and \csp\ domains algorithm
portfolios have consistently dominated the main competitions in these fields for
the past five years. For a portfolio approach to be effective there are two
crucial conditions that must be met. First, there needs to be a collection of
complementary solvers with which to make a portfolio. Second, there must be a
collection of problem features that can accurately identify structural
differences between instances. This paper focuses on the latter issue: feature
representation, because, unlike \sat, not every problem has well-studied features. We employ the well-known \satzilla\ feature set, but compute
alternative sets on different \sat\ encodings of \csp s. We show that regardless of what encoding is used to convert the instances, adequate structural information is maintained to differentiate between problem instances, and that this can be exploited to make an effective portfolio-based \csp\ solver.

\end{abstract}

\section{Introduction}
Significant strides have recently been made in the application of
portfolio-based algorithms in the fields of constraint
satisfaction~\cite{OMahony:2008vs}, quantified boolean
formulae~\cite{DBLP:conf/cp/Pulina:2007}, and most notably in
\sat~\cite{Xu:2008:SA,DBLP:conf/cp/Kadioglu:2011}. Having a collection of
solvers, these approaches compute a set of representative features about a
problem instance and then use this information to decide what is the most
effective solver to employ. These decisions can be made based on regression
techniques~\cite{Xu:2008:SA}, in which a classifier is trained to predict
expected runtime of each solver and choosing the one with best predicted
performance. Alternatively, a ranking algorithm can be trained to directly
predict the best solver for each instance~\cite{DBLP:conf/iccbr/HurleyO12}. The
features can also be used for clustering~\cite{DBLP:conf/ecai/KadiogluMST10},
where the best solver is chosen for each cluster of instances. In practice,
regardless of the approach, portfolio algorithms have been shown to be
dramatically better than using a single solver.

\smallskip
Algorithm portfolios also rely on a good set of features to describe the problem
instance being solved. This can be seen as a major drawback since one needs to use specific features for each problem at hand, or worse, has to come up with a set of features if none exists.
If there are not enough informative features present, it is impossible to train a
classifier to differentiate between classes of instances. On the other hand, if
there are too many features it is possible to over fit the classifier to the
training data. Furthermore, a large feature set is likely to have noisy
features, which could be detrimental to the quality of the learned classifier.
In the \sat\ domain, the features used by the solvers dominating the
competitions have been thoroughly analyzed and studied over the last decade.
Unfortunately, many other fields do not have such well established feature set. Even
in the case of constraint satisfaction problems, where a feature set has been
proposed, careful filtering can dramatically improve the quality of
portfolios~\cite{DBLP:conf/ictai/KroerM11}.

However, while there might not be an existing feature set, for NP-complete
problems there exist polynomial-time transformations to any other NP-complete
problem. In this paper we propose to take advantage of this by transforming \csp\
instances to \sat\ as a pre-processing step before computing its features. We
show that such a transformation retains the necessary information needed to
differentiate the classes of instances. In particular we show the effectiveness
of this approach on constraint satisfaction problems. We choose the \csp\ domain
for two reasons. First, it has a large number of solvers that can be used to
make a diversified portfolio. Second, because a feature set exists for \csp s,
we can compare the quality of a portfolio trained on \sat\ features to the
domain specific \csp\ features.

There has been a lot of work exploring the effect of transforming \csp\
instances into \sat. Perhaps the most relevant work is by Ans{\'o}tegui and
Many{\`a} which evaluated the performance of \sat\ solvers on six \sat-encodings
on graph colouring, random binary \csp s, pigeon hole, and all interval series
problems~\cite{DBLP:conf/sat/AnsoteguiM04}. Solvers such as
sugar~\cite{TamuraTB:CSC:2008}, azucar~\cite{TanjoTB:SAT:2012}, and
\texttt{CSP4SAT4J}~\cite{LeBerreLynce:CP08} have similarly tackled \csp\
problems by encoding them into \sat\ and then solving them with a predefined
\sat\ solver. However, as far as we are aware, this paper represents the first
time that  a portfolio has been created using features gained \emph{after} transforming
a problem from one domain to another.


\section{Encodings}
There are a number of known polynomial-time transformations, or encodings, from constraint
satisfaction problems to \sat~\cite{DBLP:series/faia/Prestwich09}. In this paper
we focus on three commonly used encodings: the direct, order and support encodings.


\subsection{Direct Encoding}

In the direct encoding~\cite{Walsh:2000} for each \csp\ variable $X$,
 with domain $\{1, \ldots, d\}$, a \sat\ variable 
 is created for each domain value, i.e. $x_1, x_2, \ldots, x_d$.
If $x_1$ is $true$ in the resulting \sat\
formula, then the \csp\ variable $X$ is assigned the value $1$
 in the \csp\ solution. Therefore, in order to represent a
solution to the \csp\, exactly one of $x_1, x_2, \ldots, x_d$ must be assigned
$true$. We add an \emph{at-least-one} clause and \emph{at-most-one} clauses to
the \sat\ formula for each \csp\ variable $X$:

\medskip
\noindent\begin{minipage}{0.6\textwidth}
\begin{flushright}
$(x_1 \vee x_2 \vee \ldots \vee x_d)$ \\
$\forall{v, w \in \text{D}(X)}: (\neg x_v \vee \neg x_w)$
\end{flushright}
\end{minipage}
\begin{minipage}{0.3\textwidth}
\begin{flushright}
    At Least One\\
    At Most One
\end{flushright}
\end{minipage}
\vspace{5pt}

\medskip

Constraints between \csp\ variables are represented in the direct encoding by
enumerating the conflicting tuples. For a binary constraint between the pair of
variables $X$ and $Y$, if the tuple $\langle X=v, Y=w \rangle$ is forbidden,
then we add the conflict clause $(\neg x_v \vee \neg y_w)$.

\subsection{Support Encoding}

The support encoding~\cite{Kasif:1990,DBLP:conf/ecai/Gent02} uses the same mechanism as the
direct encoding to translate a \csp\ variable's domain into \sat. However, the
support encoding differs on how the constraints between variables are encoded.
Given a constraint between two variables $X$ and $Y$, for each value $v$ in the
domain of $X$, let $S_{Y,X=v} \subset D(Y)$ be the subset of the values in the
domain of $Y$ which are consistent with assigning $X=v$. Either $x_v$ is \false\
or one of the consistent assignments from $y_1 \ldots y_d$  must be true,
represented by the clause:
\[
	\neg x_v \vee \left( \bigvee_{i \in S_{Y,X=v}} y_i \right)
\]
This must be repeated by adding clauses for each value in the domain of $Y$ and
listing the values in $X$ which are consistent with each assignment.

\subsection{Order Encoding}

Unlike the direct and support encoding which model $X=v$ as a \sat\ variable,
the order encoding creates \sat\ variables to represent $X \leq v$. If $X$ is
less than or equal to $v$, then $X$ must also be less than or equal to $v+1$. To
enforce this across the domain we add the clauses:
\[
    \forall_{v}^{d-1} : (\neg x_{\leq v} \vee x_{\leq v+1})
\]

The order encoding is naturally suited to modelling inequality constraints. To
state $X \leq 3$, we would just post the unit clause $(x_{\leq 3})$. If we want to
model the constraint $X=v$, we could rewrite it as $(X \leq v \wedge X \geq v)$.
$X \geq v$ can then be rewritten as $\neg X \leq (v-1)$. To state that $X=v$ under
the order encoding, we would encode $(x_{\leq v} \wedge \neg x_{\leq v-1})$. A
conflicting tuple between two variables, for example $\langle X=v, Y=w \rangle$
can be written in propositional logic and simplified to a \cnf\ clause
using De Morgan's Law:
\begin{align*}
\neg ((x_{\leq v} \wedge x_{\geq v}) &\wedge
      (y_{\leq w} \wedge y_{\geq w})) \\
\neg ((x_{\leq v} \wedge \neg x_{\leq v-1}) &\wedge 
      (y_{\leq w} \wedge \neg y_{\leq w-1}) ) \\
\neg (x_{\leq v} \wedge \neg x_{\leq v-1}) &\vee
\neg (y_{\leq w} \wedge \neg y_{\leq w-1}) \\
(\neg x_{\leq v} \vee x_{\leq v-1} &\vee \neg y_{\leq w} \vee y_{\leq w-1})
\end{align*}


\section{Feature Computation}

In addition to the pure direct, support and order encodings discussed in the previous section, we also consider variants of these encodings in which the clauses that encode the domains
 of the variables are not included.
We omit the domains in order to test whether focusing only on the constraints present in a
\csp\ is enough to differentiate the instances. We now briefly describe the features used for CSP and SAT. 

\medskip
{\noindent \textbf{CSP Features.}}
 We compute features for each of the original \csp\ instances, plus for each of the six encodings. We record 36 features directly from the \csp\ instance using \mistral~\cite{mistral}. This includes static features such as statistics about the types of constraints used, average and maximum domain size; and dynamic statistics recorded by running \mistral\ for 2 seconds: average and standard
deviation of variable weights, number of nodes, number of propagations and a few
others. 

\medskip
{\noindent \textbf{SAT Features.}} We use the 54 features computed using the newest feature computation tool from UBC~\cite{SATfeatures}. These features include problem size features, graph-based features, balance features, proximity to horn formula features, DPLL probing features, and local search probing features.

\section{Numerical Results}

We implemented a tool to translate a \csp\ instance
specified in XCSP format~\cite{DBLP:journals/corr/abs-0902-2362} into \sat\
(\cnf). At present, it is capable of encoding inequality and binary extensional
constraints using the direct, support and order encoding.

\medskip
{\noindent \textbf{Benchmarks.}} For our evaluation, we consider \csp\ problem instances from the \csp\ solver
competition.\footnote{\csp\ solver competition instances \\
\url{http://www.cril.univ-artois.fr/~lecoutre/benchmarks.html}} Of these, we
consider the instances that contain either inequality or binary extensional
constraints. This presents a pool of 2,433 instances, containing Graph
Colouring, Random, Quasi-random, Black Hole, Quasi-group Completion, Quasi-group
With Holes, Langford, Towers of Hanoi and Pigeon Hole problems.

\medskip
{\noindent \textbf{Portfolio Approach.}} To train our portfolios we used the ISAC
methodology~\cite{DBLP:conf/ecai/KadiogluMST10} which has been shown to work
better than a regression based approaches~\cite{Malitsky:2011:NAP:2023474.2023517}. ISAC uses the computed features to cluster the instances. Then for each cluster, the best solver in the portfolio is selected. When a new instance needs to be solved, its features are computed, it is assigned to the nearest cluster, and subsequently solved using the appropriate solver.

For our \csp\ solver portfolio we used: abscon~\cite{abscon},
csp4j~\cite{LeBerreLynce:CP08}, sat4j~\cite{le2010sat4j},
pcs~\cite{DBLP:conf/aaai/VekslerS10}, gecode~\cite{gecode}, and
sugar~\cite{TamuraTB:CSC:2008}. Each instance was run for 3,600 seconds. It is important to note that we include the time required for encoding the instances and computing the features as part of the computation time.

\begin{table}[tb]

   \caption{Comparison of number of solved instances and PAR 10 score between the virtual best solver (VBS), the portfolio approach, the random cluster approach and the best single solver based on \csp\ and \sat\ features for clustering.}

   \label{tab:results}

    \centering
    \setlength{\tabcolsep}{1ex}
    \begin{tabular}{lrrrrrrr}
        \toprule
        \multicolumn{8}{c}{PAR 10} \\ \midrule
        \multirow{2}{*}{Approach}           & CSP       & Direct        & Direct        & Order     & Order         & Support       & Support   \\
                            &           &           & \multicolumn{1}{c}{ND}        &           & \multicolumn{1}{c}{ND}        &           & \multicolumn{1}{c}{ND}        \\  \midrule
        VBS             & 1792      & 1887      & 1793      & 1806      & 1806      & 1810      & 1811      \\
        Portfolio           & 2066      & 3312      & 3221      & 2689      & 2077      & 2084      & \textbf{2022}      \\
        Random Cluster    & 3806      & 3705      & 3424      & 3725      & 3797      & 3867      & 3902      \\
        Best Single     & 4776      & 4870      & 4777      & 4789      & 4789      & 4792      & 4792      \\
        \bottomrule
        \toprule
        \multicolumn{8}{c}{Number Solved} \\ \midrule
        \multirow{2}{*}{Approach}           & CSP       & Direct        & Direct        & Order     & Order         & Support       & Support   \\
                        &           &           & \multicolumn{1}{c}{ND}        &           & \multicolumn{1}{c}{ND}        &           & \multicolumn{1}{c}{ND}        \\  \midrule
        VBS               & 2315      & 2310      & 2315      & 2315      & 2315      & 2315      & 2315      \\
        Portfolio           & 2297      & 2215      & 2220      & 2256      & 2297      & 2297      & \textbf{2301}      \\
        Random Cluster    & 2180      & 2188      & 2206      & 2187      & 2182      & 2177      & 2175      \\
        Best Single     & 2115      & 2110      & 2115      & 2115      & 2115      & 2115      & 2115      \\
        
        \bottomrule
   \end{tabular}
\end{table}

\medskip
We perform our experiments using stratified 10-fold cross validation. In Table~\ref{tab:results}, we present the performance for both the number of solved instances and the penalized runtime average PAR 10 which counts each time-out as taking 10 times the time-out to complete for each problem representation. The SAT encodings without the variable domains are marked with ND. We compare the portfolio performance to the best single solver as well as to the oracle Virtual Best Solver (VBS) which for every instance always selects the fastest solver. As we can see, using a portfolio approach for \csp\ instances is always preferable to just choosing to run a single solver.  We also compare to a random clustering approach, which randomly groups the
instances of the test set into the same number of clusters as the portfolio
method and finds the best solver to run on each group. Note that the random
clustering is trained on the same data it is evaluated on, and further that in
practice one would not know to which cluster to assign a new instance. The
random clustering approach is included to show that the clusters found by ISAC
are indeed capturing important information about the instances. We observe this
because in all cases Portfolio is better than the Random Clustering approach.

\smallskip

Table~\ref{tab:results} also shows that regardless of the encoding we use, we
can always close at least 50\% of the performance gap between the best single
solver and the virtual best one. Furthermore, we see that if we use particularly accurate encoding, which in our case is the support encoding without domain
clauses, we can even achieve slightly better performance than using features
that have been specifically designed for the problem domain.


\section{Conclusion}

In this paper we show that it is possible to encode an instance from one problem
domain to another as a preprocessing step for feature computation. In
particular, we show that even with the overhead of converting \csp\ instances to
\sat, a \csp\ portfolio trained on well established \sat\ features can perform just
as well as if it was trained on \csp\ specific features.
These findings show
that encoding techniques can retain enough information about the original
instance to accurately differentiate different classes of instances. Our results serves as a proof of concept for an automated feature generation
approach for NP-complete problems that do not have a well studied feature
vector. We consider this as a step toward problem independent feature computation for algorithm portfolios, and we plan to analyze it further and extend its applications in the future.

\section*{Acknowledgements}

The second author was supported by Paris Kanellakis fellowship at Brown University when conducting the work contained in this document. This document reflects his opinions only and should not be interpreted, either expressed or implied, as those of his current employer.

\bibliography{satbasedfeature}
\bibliographystyle{splncs03}

\end{document}